\documentclass[10pt,twocolumn,letterpaper]{article}

\usepackage{cvpr}
\usepackage{times}
\usepackage{epsfig}
\usepackage{graphicx}
\usepackage{amsmath}
\usepackage{amssymb}
\usepackage{color}
\usepackage{bm}
\usepackage{multirow}
\usepackage{multicol}
\usepackage{xspace}
\usepackage{diagbox}
\usepackage{subfigure}
\usepackage[dvipsnames]{xcolor}
 
\usepackage{comment}
\usepackage{arydshln}
\usepackage[dvipsnames]{xcolor}
\usepackage[pagebackref=true,breaklinks=true,letterpaper=true,colorlinks,bookmarks=false]{hyperref}

\cvprfinalcopy 

\def\cvprPaperID{1098}

\definecolor{yuruncolor}{HTML}{00F9DE}

\newcommand{\descr}{SOSNet\xspace}

\newcommand{\descrP}{SOSNet+\xspace}

\newcommand{\sos}{SOS\xspace}
\newcommand{\sosFull}{Second Order Similarity\xspace}
\newcommand{\sosFullLower}{second order similarity\xspace}

\newcommand{\fos}{FOS\xspace}
\newcommand{\fosFull}{First Order Similarity\xspace}

\newcommand{\RegOurs}{SOSR\xspace}
\newcommand{\RegOursFull}{Second Order Similarity Regularization\xspace}

\newcommand{\Lm}{\mathcal{L}_{\text{FOS}}}
\newcommand{\Rs}{\mathcal{R}_{\text{SOS}}}

\newcommand{\LT}{\mathcal{L}_{\text{T}}}

\ifcvprfinal\fi
\begin{document}

\title{\descr: \RegOursFull for\\ Local Descriptor Learning}

\author{Yurun Tian\thanks{Research conducted while Yurun and Xin were interns at Scape Technologies}$^{~,1,2}$ \hspace{6pt}Xin Yu$^{3}$ \hspace{6pt}Bin Fan$^{1}$ \hspace{6pt}Fuchao Wu$^{1}$ \hspace{6pt}Huub Heijnen$^{4}$ \hspace{6pt}Vassileios Balntas$^{4}$ \\
$^1$ National Laboratory of Pattern Recognition, Institute of Automation,\\
Chinese Academy of Sciences, Beijing, China\\
$^2$University of Chinese Academy of Science, Beijing, China\\
$^3$Australian Center for Robotic Vision, Australian National University\\
$^4$Scape Technologies\\
{\tt\small \{yurun.tian,bfan,fcwu\}@nlpr.ia.ac.cn  xin.yu@anu.edu.au \{huub,vassileios\}@scape.io} }

\maketitle
\begin{abstract}
Despite the fact that \sosFull (\sos) has been used with significant success in tasks such as graph matching and clustering, it has not been exploited for learning local descriptors.
In this work, we explore the potential of \sos in the field of descriptor learning by building upon the intuition that a positive pair of matching points should exhibit similar distances with respect to other points in the embedding space. 
Thus, we propose a novel regularization term, named \RegOursFull (\RegOurs), that follows this principle. 
By incorporating \RegOurs into training, our learned descriptor achieves state-of-the-art performance on several challenging benchmarks containing distinct tasks ranging from local patch retrieval to structure from motion. 
Furthermore, by designing a von Mises-Fischer distribution based evaluation method, we link the utilization of the descriptor space to the matching performance, thus demonstrating the effectiveness of our proposed \RegOurs.
Extensive experimental results, empirical evidence, and in-depth analysis are provided, indicating that \RegOurs can significantly boost the matching performance of the learned descriptor. 
\end{abstract}

\section{Introduction}

\begin{figure}[ht]
    \centering
    \subfigure[Triplet Loss]{\includegraphics[trim={2.8cm 1cm 3.1cm 2cm},clip,width=0.23\textwidth]{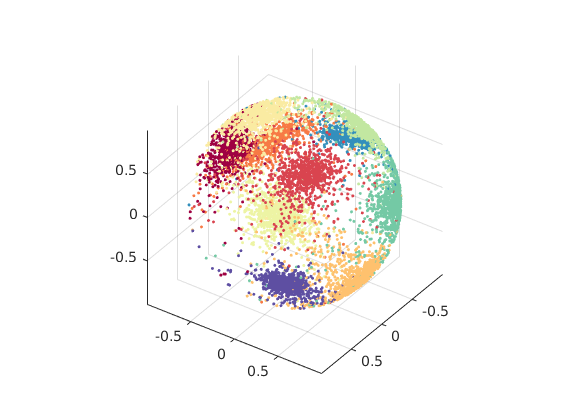}}
    \subfigure[Triplet Loss + \RegOurs]{\includegraphics[trim={2.8cm 1cm 3.1cm 2cm},clip,width=0.23\textwidth]{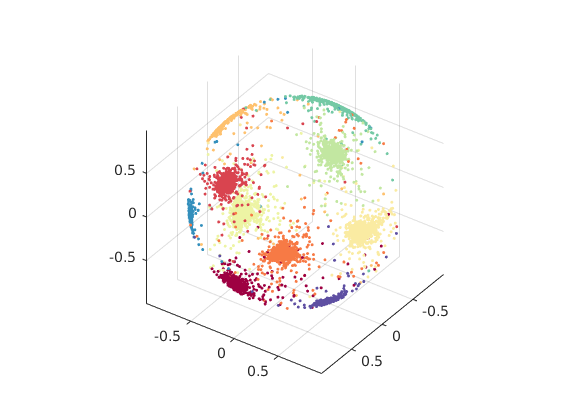}}
    \caption{Qualitative results of our proposed \RegOurs on features learned for the 10 digits of the MNIST~\cite{mnist1998} dataset. Each digit is represented by a different colour on the unit sphere. We can observe that by using our \RegOurs method that encourages second order similarity, more compact individual clusters are learned compared to standard triplet loss.}
    \label{fig:mnist_toy}
\end{figure}

The process of describing local patches is a fundamental component in many computer vision tasks such as 3D reconstruction \cite{colmapcvpr2016,colmapeccv2016}, large scale image localization \cite{sattler6dof} and image retrieval \cite{radenovic2016cnn}.
Early efforts mainly focused on the heuristic design of hand-crafted descriptors, by applying a set of filters to the input patches.
In recent years, large datasets with corresponding ground truths have led to the development of large scale learning methods, which stimulated a wave of works on descriptor learning. Recent work has shown that these learning based methods are able to significantly outperform their hand-crafted counterparts~\cite{hpatches2017,geodesc2018}.

One of the most important challenges of learning based methods is the design of suitable loss functions for the training stage. Since nearest neighbour matching is done directly using Euclidean distances, most of the recent methods focus on optimizing objectives related to \fosFull (\fos) by forcing descriptors from matching pairs to have smaller $L2$ distance than non-matching ones ~\cite{deepdesc2015,tfeat2016,hardnet2017,scaleaware2018,l2net2017,doap2018}. 

\sosFull (\sos) has been used for graph matching and clustering tasks~\cite{reweighted2010,progressivegm2012,factorizedgm}, due to the fact that it can capture more structural information such as shape and scale, while at the same time being robust to deformations and distortions. On the other hand, \fos and nearest neighbor matching are only limited to pairwise comparisons. However, utilizing \sos for large scale problems typically requires significant computational power~\cite{reweighted2010,progressivegm2012,factorizedgm}, and thus matching and reconstruction tasks still rely on brute force or approximate nearest neighbor matching \cite{colmapcvpr2016}.
In this work, we explore the possibility of using \sos for learning high performance local descriptors.
In particular, we are interested in formulating a \sos constraint as a regularization term during training, in order to harness its power during the matching stage, without any computational overhead.

Evaluation of descriptors is also a key issue.
A good evaluation method can provide insights for designing descriptors. 
Performance indicators, such as false positive rate ~\cite{ubc2011} and mean average precision~\cite{hpatches2017}, are widely used. 
However, it is still unclear how the utilization of the descriptor space, such as the degrees of intra-class concentration and inter-class dispersion, contributes to the final performance.
Therefore, in order to explain the impact of \sos on the matching performance, we further introduce an evaluation method based on the von Mises-Fisher distribution~\cite{vmf_book1981}.

Our main contributions are:
(1) We introduce a novel regularization method, named  \RegOursFull (\RegOurs), that enforces \sosFull (\sos) consistency. To the best of our knowledge, \sos has not been incorporated into the process of learning local feature descriptors.
(2) By combining our \RegOurs with a triplet loss, our learned descriptor is able to significantly outperform previous ones and achieves state-of-the-art results on several benchmarks related to local descriptors.
(3) We introduce a new evaluation method that is based on the von Mises-Fisher distribution for examining the utilization of the descriptor space. The proposed evaluation method can illustrate links between distributions of descriptors on the hypersphere and their matching performance.



This paper is organized as follows: In Sec.~\ref{sec:related_work}, we briefly review related works. In Sec.~\ref{sec:methodology} and Sec.~\ref{sec:vMF Distribution}, we introduce our \RegOursFull as well as a new method for evaluating the descriptors on a unit hypersphere.
Subsequently, in Sec.~\ref{sec:Experiments}, we present results on several challenging benchmarks. Lastly, we conduct ablation study of our proposed \RegOurs in Sec.~\ref{sec:Discussion}.

\section{Related works}\label{sec:related_work}

Early works on local patch description focused on low level processes such as gradient filters and intensity comparisons, including SIFT~\cite{sift2004},  GLOH~\cite{GLOH2005}, DAISY \cite{pickingbestdaisy2009}, DSP-SIFT~\cite{dspsift2015} and LIOP~\cite{liop2011}. A comprehensive review can be found in~\cite{mikolajczyk2005performance}.

With the emergence of annotated patch datasets~\cite{ubc2011}, a significant amount of data-driven methods focused on improving hand-crafted representations using machine learning methods.
The authors of ~\cite{discriminative2011,learningproject2011} use linear projections to learn discriminative descriptors, while convex optimization is used for learning an optimal descriptor sampling configuration in \cite{convexoptim2014}.
BinBoost~\cite{binboost2013} is trained based on a boosting framework, and in
RFD~\cite{rfd2014}, the most discriminative receptive fields are learned based on the labeled training data. 
BOLD~\cite{bold2015} uses patch specific adaptive online selection of binary intensity tests.

Convolutional neural networks (CNNs) enable end-to-end descriptor learning from raw local patches, and have become the de-facto standard for learning local patch descriptors in recent years. MatchNet~\cite{matchnet2015} adopts a Siamese network for local patch matching, while DeepCompare~\cite{deepcompare2015} further explores various network architectures.
Song~\textit{et al.}~\cite{oh2016deep} propose the lifted structured embedding for the task of feature embedding.
DeepDesc~\cite{deepdesc2015} removes the need for a specially learned distance metric layer, and instead uses Euclidean distances and hard sample mining.
TFeat~\cite{tfeat2016} uses triplet learning constraints with shallow convolutional networks and fast hard negative mining, and L2Net \cite{l2net2017} applies progressive sampling with a loss function that takes into account the whole training batch while producing descriptors that are normalized to unit norm. The L2Net architecture was widely adopted by consequent works.
HardNet \cite{hardnet2017} surpasses L2Net by implementing a simple hinge triplet loss with the ``hardest-within-batch'' mining, confirming the importance of the mining strategy. 
Keller~\textit{et al.}~\cite{scaleaware2018} propose to learn consistently scaled descriptors by a mixed context losses and a scale aware sampling.
Instead of focusing on patch matching, DOAP~\cite{doap2018} imposes a retrieval based ranking loss and achieves the current state of the art performance on several benchmarks. 
GeoDesc~\cite{geodesc2018} integrates geometry constraints from multi-view reconstructions to benefit the learning process by improving the training data. The authors of\cite{spread_out}, propose a Global Orthogonal Regularization term to better exploit the unit hypersphere.  

While the recent improvements in the field of learning CNN patch descriptors are significant, methods mentioned above are limited to optimizing the \fos measured by $L_2$ distances of positive and negative pairs, and the potential of \sos has not been exploited in the area of descriptor learning. On the other hand, 
graph matching algorithms~\cite{reweighted2010,progressivegm2012,factorizedgm} have been developed based on \sos due to its robustness to shape distortions. Furthermore, \cite{dimbless2018} proves that using \sos can achieve better cluster performance.
Thus, our key idea is to introduce \sosFullLower constraints at the training stage for robust patch description.

\section{Learning Descriptor with Second Order Similarities}\label{sec:methodology}

In this section, we introduce how to incorporate \sos as a regularization term into our training procedure. By employing \fos and \sos losses, we train our network in an end-to-end manner.

\subsection{Preliminaries}

For a training batch consisting of $N$ pairs of matching patches, a convolutional neural network is applied to each patch so as to extract its descriptor. 
The corresponding positive descriptor pairs are denoted as ${\{ {{{\bm x}_i},{\bm x}_i^ +} \}}_{{i = 1 \dots N}}$

\subsection{\fosFull Loss}
\label{sec:qsh}
\fosFull (\fos) loss, which enforces distances between matching descriptors to be small while those of non-matching ones to be large, 
has been widely used for learning local descriptors~\cite{tfeat2016,l2net2017,doap2018,geodesc2018}. 
In our method, we first employ a loss term to constraint the \fos as follows:

\begin{equation}
\label{eq:triplet_mining}
\begin{gathered}
\Lm = \frac{1}{N}\sum_{i=1}^{N} \max\left( 0, t + d^{\text{pos}}_i - d^{\text{neg}}_i\right)^2,
\\
d^{\text{pos}}_i = d(\bm{x}_i, \bm{x}_i^+),
\\
d^{\text{neg}}_i = \min\limits_{\forall j, j\neq i}(d(\bm{x}_i, \bm{x}_j), d(\bm{x}_i, \bm{x}^+_j), d(\bm{x}_i^+, \bm{x}_j), d(\bm{x}_i^+, \bm{x}^+_j)),
\end{gathered}
\end{equation}
where $t$ is the margin, $d(\bm{u},\bm{v}) = \|\bm{u}-\bm{v}\|_2$ is the $L_2$ distance, $d^{pos}_i$ indicates the distance between a positive pair \ and $d^{neg}_i$ represents the distance between a negative pair. 
We employ the same mining strategy as in HardNet~\cite{hardnet2017} to find the ``hardest-within-batch'' negatives.
Note that, in Eqn.~\eqref{eq:triplet_mining}, we use a Quadratic Hinge Triplet (QHT) loss instead of the conventional Hinge Triplet (HT) Loss.
Compared with HT, QHT weights the gradients with respect to the parameters of the network
by the magnitude of the loss.
This means that the larger $d^{\text{neg}}_i - d^{\text{pos}}_i$ is, the smaller the gradients are. 
In Sec.~\ref{sec: discuss1} we provide evidence that this simple modification can lead to significant performance improvements. 

\subsection{Second Order Similarity Regularization}
\label{sec:R_D}

Besides the first order constraints imposed by $\Lm$,
it has been demonstrated that incorporating information from higher order similarities can improve the performance of clustering~\cite{dimbless2018} and graph matching~\cite{reweighted2010}. 
Thus, we propose to impose a second order constraint to further supervise the process of descriptor learning.

A training mini-batch can be viewed as two sets of descriptors with one-to-one correspondence, \ie, $\{\bm{x}_i\}_{i = 1 \dots N}$ and $\{\bm{x}_i^+\}_{i = 1 \dots N}$. 
For this case, we  define the \sosFullLower between ${\bm{x}_i}$ and ${\bm{x}_i^+}$ as:
\begin{equation}
\label{eq:sod_def}
d^{(2)}(\bm{x}_i, \bm{x}_i^+) = \sqrt{\sum_{j \neq i}^{N}{(d(\bm{x}_i, \bm{x}_j) - d(\bm{x}^+_i, \bm{x}^+_j))^2}},
\end{equation}
where  $d^{(2)}(\bm{x}_i, \bm{x}_i^+)$ measures the similarity between $\bm x_i$ and $\bm x_i^{+}$ from the perspectives of $\{\bm x_j\}_{j \neq i}$ and $\{\bm x_j^{+}\}_{j \neq i}$, using the differences between distances.

In order to enforce \sos, we formulate our \RegOurs regularization term as:
\begin{equation}
\label{eq:sod_objective}
\Rs = \frac{1}{N}\sum_{i=1}^{N}d^{(2)}(\bm{x}_i, \bm{x}_i^+).
\end{equation}

Note that $\Rs$  does not force distances between matching descriptors to decrease or distances between non-matching ones to increase. Thus, it cannot be solely used without an $\Lm$ term, and can only be served as a regularization term.

\subsection{Objective Function For Training}

Our goal is to learn a robust descriptor in terms of both \fos and \sos, therefore, our total objective function is expressed as:
\begin{equation}
\label{eq:objective}
\LT = \Lm+\Rs, 
\end{equation}
where the two terms are weighted equally. 

\subsection{Implementation Details}

During training, we observed that using all the samples in a mini-batch as input to the $\Rs$ term led to inferior results. This is due to the fact that for a given pair of matching descriptors, many of their non-matching descriptors are already far away.
Thus, these distant negatives need no further optimization.
Subsequently, \sos calculated on these ``easy'' negatives may produce noisy gradients and therefore damage the performance.
Inspired by the concept of active graph in~\cite{progressivegm2012}, we employ nearest neighbor search to exclude those far away negatives for each positive pair.
Let $z_i$ be the class label of the $i^{\text{th}}$ positive pair, and $\bm{c}_i$ be the $i^{\text{th}}$ set of class labels. In particular, $\bm{c}_i$ stores the class labels which are within the $K$ Nearest Neighbors ($K$NN) of the $i^{\text{th}}$ positive pair. Thus, we define the criterion of neighbor selection for each $\bm{c}_i$ as:
\begin{equation}
\label{eq:c_i}
\begin{gathered}
\bm{c}_i = \{z_j:  \bm{x}_j \in K{\rm NN}(\bm{x}_i) \vee \bm{x}_j^+ \in K{\rm NN}(\bm{x}_i^+)\}, 
\\
\forall j \in 1 \dots N, j \neq i
\end{gathered}
\end{equation}
where $K{\rm NN}(\bm{x}_i)$ denotes the $K$ Nearest Neighbors of descriptor $\bm{x}_i$. 
Note that there is a possibility of intersection between the $K${\rm NN}($\bm{x}_i$) and $K${\rm NN}($\bm{x}_i^+$) sets. Thus, the cardinality of $\bm{c}_i$ ranges from $K$ to $2K$. Therefore, in Eqn~\eqref{eq:sod_objective} we calculate \sos for the $i^{\text{th}}$ pair as:
\begin{equation}
\label{eq:sod_knn}
d^{(2)}(\bm{x}_i, \bm{x}_i^+) = \sqrt{\sum_{j \neq i,z_j \subset \bm{c}_i}^{N}{(d(\bm{x}_i, \bm{x}_j) - d(\bm{x}^+_i, \bm{x}^+_j))^2}}.
\end{equation}

We adopt the architecture of L2Net~\cite{l2net2017} to embed local patches to $128$-dimensional descriptors.  Note that all descriptors are normalized to unit vectors. 
To prevent overfitting, we also employ a dropout layer with a drop rate of $0.1$ before the last convolutional layer. Similar to previous works~\cite{l2net2017, hardnet2017}, all patches are resized to $32 \times 32$ and normalized by subtracting the per-patch mean and dividing the per-patch standard deviation. We use the PyTorch library \cite{pytorch2017} to train our local descriptor network.
Our network is trained for 100 epochs using the Adam optimizer \cite{adam2014} with $\alpha =0.01$, $\beta_1 =0.9$ and $\beta_2 =0.999$ as in the default settings. For the training hyperparameters, the number of training pairs $N$ is set to $512$, \ie, the batch size is $1024$, 
$K$ is set to $8$, that is, $8$ nearest neighboring pairs are selected to calculate \sos for a given pair, and the margin $t$ in the \fos loss is set to be $1$.

\section{Evaluating the Unit Hypersphere Utilization}
\label{sec:vMF Distribution}

Indicators like false positive rate and mean average precision have been widely used for evaluating the performance of descriptors \cite{hpatches2017,pickingbestdaisy2009}.
However, such indicators fail to provide insights on properties of the learned descriptors, \ie, how the utilization of the descriptor space such as the intra-class and inter-class distributions, contribute to the final performance.
To investigate this, previous works\cite{spread_out,l2net2017} visualize the distributions of the positive and negative distances as histograms. However, while such visualizations illuminate the distance distributions, they fail to capture the structure of the learned descriptor space.

Since most modern methods rely on normalized descriptors, we propose to leverage the von Mises-Fisher (vMF) distribution which deals with the statistical properties of unit vectors that lie on hyperspheres (interested readers can find more information in \cite{vmf_book1981}). A $q$-dimensional descriptor, can be thought as a random point on the $(q-1)$-dimensional unit hypersphere $\mathbb{S}^{q-1}$. 
Specifically, a random unit vector $\bm{x}$ (\ie, $\|\bm{x}\|_2 = 1$) obeys a $q$-variate vMF distribution if its probability density function is as follows: 
\begin{equation}\label{eq:vMF_def}
f(\bm{x} | \bm{\mu}, \kappa) = c_q(\kappa)e^{\kappa \bm{\mu}^{\rm{T}} \bm{x}},
\end{equation}
where $\|\bm{\mu}\|_2 = 1$, $q \geq 2$ and $\kappa \geq 0$.
The normalizing constant $c_q(\kappa)$ is defined as:
\begin{equation}\label{eq:vMF_const}
c_q(\kappa) = \frac{\kappa^{q/2-1}}{(2\pi)^{q/2}I_{q/2-1}(\kappa)},
\end{equation}
where $I_{k}(\cdot)$ is the modified Bessel function of the first kind and order $k$. 
The vMF density $f(\bm{x} | \bm{\mu}, \kappa)$ is parameterized by the mean direction $\bm{\mu}$ and the concentration parameter $\kappa$. $\kappa$ is used to characterize how strongly the unit vectors drawn from $f(\bm{x} | \bm{\mu}, \kappa)$ are concentrated in the mean direction $\bm{\mu}$, with larger values of $\kappa$ indicating stronger concentration.
In particular, when $\kappa = 0$, $f(\bm{x} | \bm{\mu}, \kappa)$ reduces to the uniform distribution on $\mathbb{S}^{q-1}$, and as $\kappa \rightarrow \infty$, $f(\bm{x} | \bm{\mu}, \kappa)$ approaches a point density.

According to~\cite{vmf_book1981}, the maximum likelihood estimation of $\kappa$ can be obtained from the following equation:
\begin{equation}\label{eq:def_R}
A(\hat{\kappa}) = \frac{I_{q/2}(\hat{\kappa})}{I_{q/2-1}(\hat{\kappa})}  = \Bar{R} = \frac{1}{N}\|\sum_{i=1}^{N}\bm{x}_i\|_2,
\end{equation}
where $\hat{\kappa}$ is the estimation of $\kappa$ and $\Bar{R}$ is called the mean resultant length. 
Since $A(\cdot)$ is a ratio of Bessel functions \cite{vMFclustering} with no analytic inverse, we cannot directly take $\hat{\kappa} = A^{-1}(\Bar{R})$.
According to \cite{vmfclustering2005}, $\hat{\kappa}$ can be approximated by a monotonically increasing function of $\Bar{R}$, where $\Bar{R}=0$ leads to $\hat{\kappa}=0$ and $\Bar{R}=1$ indicates $\hat{\kappa}=\infty$.
Therefore, $\Bar{R}$ can be used as a proxy for measuring $\kappa$.

Descriptors from the $i^{\text{th}}$ class can be interpreted as samples drawn from a vMF distribution $f^i_{\text{intra}}(\bm{x} | \bm{\mu}_i, \kappa^i_{\text{intra}})$.  The cluster center $\bm{\mu}_i$ is a sample from vMF density $f_{\text{inter}}(\bm{\mu} | \bm{\nu}, \kappa_{\text{inter}})$.
Further, to investigate the utilization of the unit hypersphere, we define the following parameters:
\begin{equation}
\label{eq:vmf_indicator}
\begin{gathered}
R_{\text{intra}} = \frac{1}{M}\sum_{i}^{M}{\bar{R}^{\text{intra}}_{i}},\\
R_{\text{inter}} = \frac{1}{M}{\|\sum_{i=1}^{M}\bm{\mu}_i\|_2},
\\
\rho = \frac{R_{\text{inter}}}{R_{\text{intra}}},
\end{gathered}
\end{equation}
where $M$ is the total number of classes and $\bar{R}^{\text{intra}}_{i}$ is the mean resultant length for the $i^{\text{th}}$ class. 
In Eqn.~\eqref{eq:vmf_indicator}, $R_{\text{intra}}$ and $R_{\text{inter}}$ measure the intra-class concentration, inter-class dispersion respectively, and the ratio $\rho$ is a overall evaluation.

The vMF distribution has been used in image clustering~\cite{vMFclustering} and  classification~\cite{vmfclassification2018}. 
However, we propose to use it solely for evaluating the utilization of the descriptor space, since unlike in the classification tasks, current local patch datasets can not guarantee sufficient intra-class samples for accurate estimation of the {\em vMF parameters}, \eg, some classes in the widely used UBC Phototour~\cite{ubc2011} dataset only have $2$ samples, and such estimation errors in training stage may lead to inferior performance.

\begin{table*}[htp]
\begin{center}
\begin{tabular}{c| c c c c c c c c c c}
\hline
Train& Notredame & Yosemite && Liberty & Yosemite &&Liberty &Notredame & \multirow{2}*{Mean}\\
\cline{1-1} \cline{2-3} \cline{5-6}  \cline{8-9}

Test&\multicolumn{2}{c}{Liberty}&& \multicolumn{2}{c}{Notredame}&&\multicolumn{2}{c}{Yosemite}& \\ \hline

SIFT \cite{sift2004}&\multicolumn{2}{c}{29.84}&&\multicolumn{2}{c}{22.53}& &\multicolumn{2}{c}{27.29} &26.55\\
DeepDesc \cite{deepdesc2015}     	&\multicolumn{2}{c}{10.9}   &&   \multicolumn{2}{c}{4.40} && \multicolumn{2}{c}{5.69} &6.99\\

MatchNet \cite{tfeat2016}   &7.04 &11.47  &&3.82&5.65&&11.6&8.70 &8.05\\
L2Net \cite{l2net2017}  &3.64&5.29 &&1.15&1.62&&4.43&3.30&3.24\\
CS L2Net \cite{l2net2017}  &2.55&4.24 &&0.87&1.39&&3.81&2.84&2.61\\
HardNet \cite{hardnet2017} &1.47&\bf{2.67} &&0.62&0.88&&2.14&1.65&1.57\\
HardNet-GOR~\cite{hardnet2017,spread_out} &1.72&2.89&&0.63&0.91&&2.10&1.59&1.64\\
Michel \textit{et al}.~\cite{scaleaware2018} &1.79&2.96 &&0.68&1.02&&2.51&1.64&1.77\\
\descr &\bf{1.25}&2.84 &&\bf{0.58}&\bf{0.87}&&\bf{1.95}&\bf{1.25}&\bf{1.46}\\
\hline
TFeat+~\cite{tfeat2016}   &7.39 &10.13  &&3.06&3.80&&8.06&7.24 &6.64\\
L2Net+~\cite{l2net2017} &2.36&4.70 &&0.72&1.29&&2.57&1.71&2.23\\
CS L2Net+~\cite{l2net2017}  &1.71&3.87 &&0.56&1.09&&2.07&1.3&1.76\\
HardNet+~\cite{hardnet2017} &1.49&2.51&&0.53&0.78&&1.96&1.84&1.51\\
HardNet-GOR+~\cite{hardnet2017,spread_out}  &1.48&2.43&&0.51&0.78&&1.76&1.53&1.41\\
DOAP+ \cite{doap2018}  &1.54&2.62&&0.43&0.87&&2.00&1.21&1.45\\
DOAP-ST+ \cite{doap2018} \cite{stn2015}  &1.47&2.29&&0.39&0.78&&1.98&1.35&1.38\\
\descrP &\bf{1.08}&\bf{2.12} &&\bf{0.35}&\bf{0.67}&&\bf{1.03}&\bf{0.95}&\bf{1.03}\\
\hdashline
GeoDesc+~\cite{geodesc2018}     	&\multicolumn{2}{c}{5.47}   &&   \multicolumn{2}{c}{1.94} && \multicolumn{2}{c}{4.72} &4.05\\
\descr-HP+   &\multicolumn{2}{c}{2.10}   &&   \multicolumn{2}{c}{0.79} && \multicolumn{2}{c}{1.39} &1.42\\
\hline
\end{tabular}
\end{center}
\caption{Patch verification performance on the UBC phototour dataset. Numbers denote false positive rates at 95\% recall. 
All descriptors are $128$-dimensional, except that TFeat is $256$. and suffix ``+'' indicates data augmentation. We can observe that our \descr outperforms other methods in all cases. }
\label{tab:UBC_performance}
\end{table*}

\section{Experiments}
\label{sec:Experiments}

We name our learned descriptor {\em Second Order Similarity Network} (\descr). 
In this section, we compare our \descr with several state-of-the-art methods, \ie, DeepDesc(DDesc)~\cite{deepdesc2015}, TFeat~\cite{tfeat2016}, L2Net~\cite{l2net2017}, HardNet(HNet)~\cite{hardnet2017}, HardNet with GOR~\cite{spread_out}, Scale-Aware Descriptor~\cite{scaleaware2018}, DOAP~\cite{doap2018} and GeoDesc~\cite{geodesc2018}.
We perform our experiments on three publicly available datasets, namely UBC Phototour \cite{ubc2011}, HPatches \cite{hpatches2017} and ETH SfM\cite{eth_benchmark2017}.
For TFeat \cite{tfeat2016}, L2Net \cite{l2net2017}, HardNet \cite{hardnet2017}, and GeoDesc\footnote{The training dataset of GeoDesc is not publicly available. Therefore, comparisons of GeoDesc with other methods may be unfair.}~\cite{geodesc2018}, we use the pre-trained models released by the authors, and for GOR \cite{spread_out}, we employ the code provided by the authors.
For Scale-Aware Descriptors~\cite{scaleaware2018} and DOAP \cite{doap2018}, we report their results from their published papers since their training codes and pre-trained models are unavailable.

\subsection{UBC Phototour}

UBC Phototour dataset \cite{ubc2011} is currently the most widely used dataset for local patch descriptor learning.
It consists of three subsets, Liberty, Notredame, and Yosemite.
For evaluations on this dataset, models are trained on one subset and tested on the other two. 
We follow the standard evaluation protocol of \cite{ubc2011} by using the 100K pairs provided by the authors and report the false positive rate at 95\% recall.

In Table.~\ref{tab:UBC_performance}, SIFT represents a baseline hand-crafted descriptor, and the others are CNN based methods.
As indicated by Table.~\ref{tab:UBC_performance}, our \descr achieves the best performance by a significant margin in comparison to the state-of-the-art approaches.
Note that, DOAP incorporates a Spatial Transformer Network (STN)~\cite{stn2015} into the network to resist geometrical distortions in the patches. In contrast, our \descr does not require any extra geometry rectifying layer and yet achieves superior performance. We can expect the performance of our method to further increase, by incorporating an STN.
GeoDesc generates inferior results due to the possible differences between its training dataset and UBC Phototour.
In addition, it is worth noting that even when trained on HPatches (\descr-HP+), our descriptor is able to closely match the best performing methods, which is significant since UBC and HPatches exhibit vastly different patch distributions. This is a testament to the generalization ability of our method.

\begin{figure*}[ht]
\centering
\includegraphics[trim={0cm 0cm 0cm 0cm}, width=\linewidth]{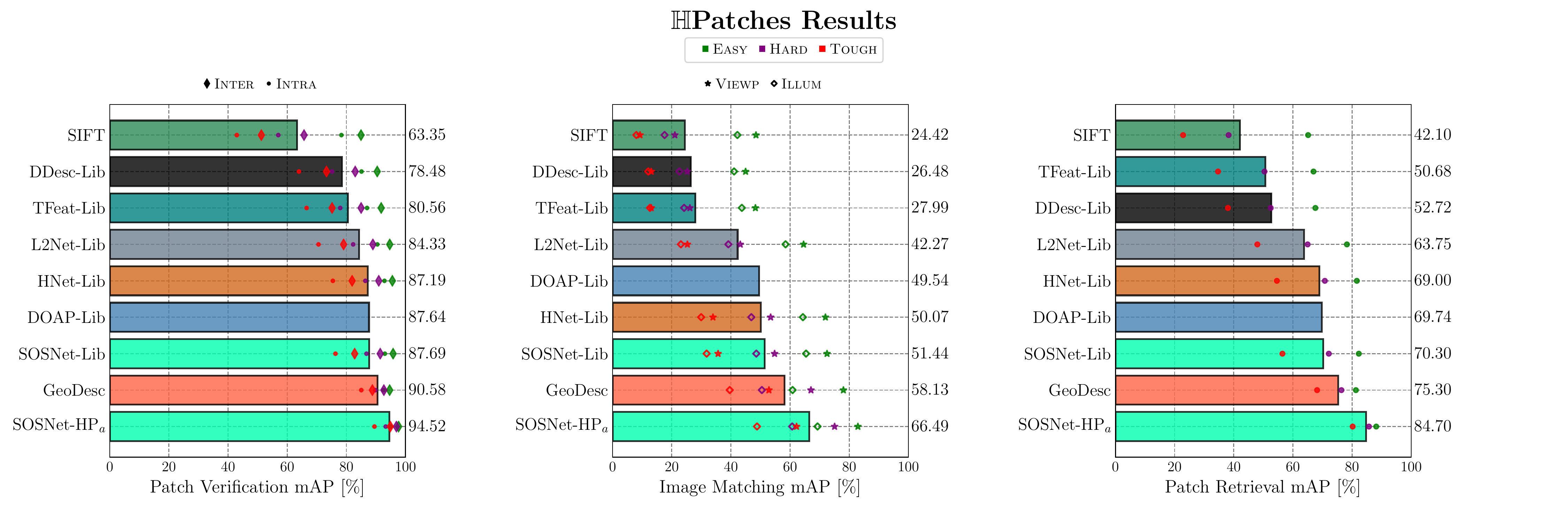}
\caption{Verification, matching and retrieval results on test set split `a' of HPatches~\cite{hpatches2017}. Colour of the marker indicates EASY, HARD, and TOUGH noise. The type of the marker corresponds to the variants of the experimental settings. }
\label{fig:hpatehes_results}
\end{figure*}

\subsection{HPatches}
\label{sec:HPatches}

HPatches dataset~\cite{hpatches2017} consists of over 1.5 million patches extracted from 116 viewpoint and illumination changing scenes. According to the different levels of geometric noise, the extracted patches can be divided into three groups: {\em easy}, {\em hard}, and {\em tough}. There are three evaluation tasks, patch verification, patch retrieval, and image matching. We show results for all three tasks in Fig.~\ref{fig:hpatehes_results}.
As shown in Fig.~\ref{fig:hpatehes_results}, our \descr outperforms state-of-the-art methods on all the three tasks both for methods trained on Liberty (\texttt{-LIB}) and on HPatches (\texttt{-HP}). It is worth noting that our descriptor outperforms DOAP in the retrieval task, even though DOAP employs a ranking loss specifically designed for maximizing the mean Average Precision (mAP) for patch retrieval. This indicates that our \RegOurs can lead to more discriminative descriptors, without the need for a specialized ranking loss.

\subsection{ETH dataset}

Different from the aforementioned datasets that focus on patches, the ETH SfM benchmark \cite{eth_benchmark2017} aims to evaluate descriptors for a Structure from Motion (SfM) task. This benchmark investigates how different methods perform in terms of building a 3D model from a set of available 2D images.  

In this experiment, we compare our \descr with the state-of-the-art methods by quantifying the SfM quality, \ie, measuring the number of registered images, reconstructed sparse points, image observations, mean track length and mean reprojection error. Following the protocols in \cite{eth_benchmark2017}, we do not conduct the ratio test, in order to investigate the direct matching performance of the descriptors.

Table~\ref{tab:eth_results} shows the evaluation results of the 3D reconstruction, in which \descr exhibits the best overall performance. In particular, \descr is able to significantly outperform other methods in terms of metrics related to the density of the reconstructed 3D model, \ie the number of registered sparse points, and the number of observations. It is worth noting that
\descr produces even more matches than GeoDesc~\cite{geodesc2018}, which is specifically designed and trained for a SfM task.
Similarly to the observations in~\cite{eth_benchmark2017,geodesc2018}, SIFT achieves the smallest reprojection error on all tests, thus demonstrating that it is still an attractive choice for image matching. This can be explained by the fact that fewer matches seem to lead to a trend for lower reprojection errors. Furthermore, since the reprojection errors are less than $1$px for all descriptors, we can conclude that this metric may not reflect performance differences between descriptors in practice. Finally, we can observe that our method is able to register significantly more images compared to SIFT. For example, in  \texttt{Madrid Metropolis} sequence, SIFT was able to register only 38\% of the available 2D images for the final 3D model, while our method registered 65\% of the images. This indicates that our method is more suitable for large scale and challenging reconstructions. 

\section{Discussion}
\label{sec:Discussion}
In this section, we perform several experiments to provide a more in-depth analysis about how each component in \descr contributes to its final performance. 
Besides reporting matching performance in terms of FPR@95 rate and mAP, we also demonstrate how the proposed \RegOurs and other existing methods impact on the structure of the learned descriptor space, using the methodology introduced in Sec.~\ref{sec:vMF Distribution}.

\begin{table*}[htp]
\footnotesize
\begin{center}
\begin{tabular}{c l c c c c c c}
\hline

& & \bf{\# Image} & \bf{\# Registered} 
& \bf{\# Sparse Points} & \bf{\# Observations} & \bf{Track Length} & \bf{Reproj. Error}  \\
\hline  
\bf{Fountain} & SIFT & 11 & 11 & 14K & 70K & 4.79 & 0.39px\\
& DSP-SIFT & & 11  & 14K & 71K & 4.78 & \textcolor{red}{0.37px}\\
& L2Net & & 11  & 17K & 83K & 4.88& 0.47px \\
& GeoDesc & & 11  & 16K & 83K& \textcolor{red}{5.00}& 0.47px \\
& \descr & & 11 & \textcolor{red}{17K} & \textcolor{red}{85K} & 4.92& 0.43px\\

\hline

\bf{Herzjesu} & SIFT & 8 & 8 & 7.5K & 31K & 4.22 & \textcolor{red}{0.43px}\\
& DSP-SIFT & & 8 & 7.7K & 32K & 4.22 & 0.45px\\
& L2Net & & 8 & 9.5K & 40K & 4.24 & 0.51px\\
& Geodesc & & 8 & 9.2K & 40K & \textcolor{red}{4.35} & 0.51px\\
& \descr & & 8 & \textcolor{red}{9.7K} & \textcolor{red}{41K} & 4.26 & 0.53px\\
\hline

\bf{South Building} & SIFT & 128 & 128 & 108K & 653K & \textcolor{red}{6.04} & \textcolor{red}{0.54px}\\
& DSP-SIFT & & 128 & 112K & 666K & 5.91 & 0.58px\\
& L2Net & & 128 &170K & 863K& 5.07& 0.63px\\
& GeoDesc & & 128 & 170K & 887K & 5.21 & 0.64px\\
& \descr & & 128 & \textcolor{red}{178K} & \textcolor{red}{913K} & 5.11 & 0.67px\\
\hline

\bf{Madrid Metropolis} & SIFT & 1344 & 500 & 116K & 733K & 6.32 & \textcolor{red}{0.60px}\\
& DSP-SIFT & & 467 & 99K & 649K & \textcolor{red}{6.52} & 0.66px\\
& L2Net & & 692 & 254k & 1067K & 4.20 & 0.69px \\

& GeoDesc &  &809 & 306K & 1200K & 3.91 & 0.66px\\

& \descr & & \textcolor{red}{844}  & \textcolor{red}{335K} & \textcolor{red}{1411K} & 4.21 & 0.70px\\
\hline
\bf{Gendarmenmarkt} & SIFT & 1463 & 1035  &338K& 1872K & \textcolor{red}{5.523} & \textcolor{red}{0.69px}\\
& DSP-SIFT & & 979  & 293K &1577K & 5.381  &0.74px\\
& L2Net & & 1168& 667k & 2611K & 3.91 & 0.73px\\

& GeoDesc &  & \textcolor{red}{1208}& 779K & 2903K & 3.72 & 0.74px\\
& \descr & & 1201  & \textcolor{red}{816K} & \textcolor{red}{3255K} & 3.984 & 0.77px\\

\hline
\end{tabular}
\end{center}
\normalsize
\caption{Evaluation results on ETH dataset \cite{eth_benchmark2017} for SfM. We can observe that our proposed \descr significantly outperforms other methods in terms of the number of registered sparse points and number of observations. This indicates that the models that are built using our descriptor are significantly denser.}
\label{tab:eth_results}
\end{table*}

\subsection{Analysis of Performance Improvements}
\label{sec: discuss1}

We argue that the performance increase of \descr comes from three aspects: 1) the optimization method that we employ, 2) the QHT, and 3) the proposed \RegOurs.

\begin{figure*}
\centering
\begin{minipage}{0.325\linewidth}
\subfigure[Training with SGD.]{\label{fig:FRR_epoch_sgd}\includegraphics[trim={1.5cm 0.5cm 1.5cm 1.2cm},width=1\linewidth]{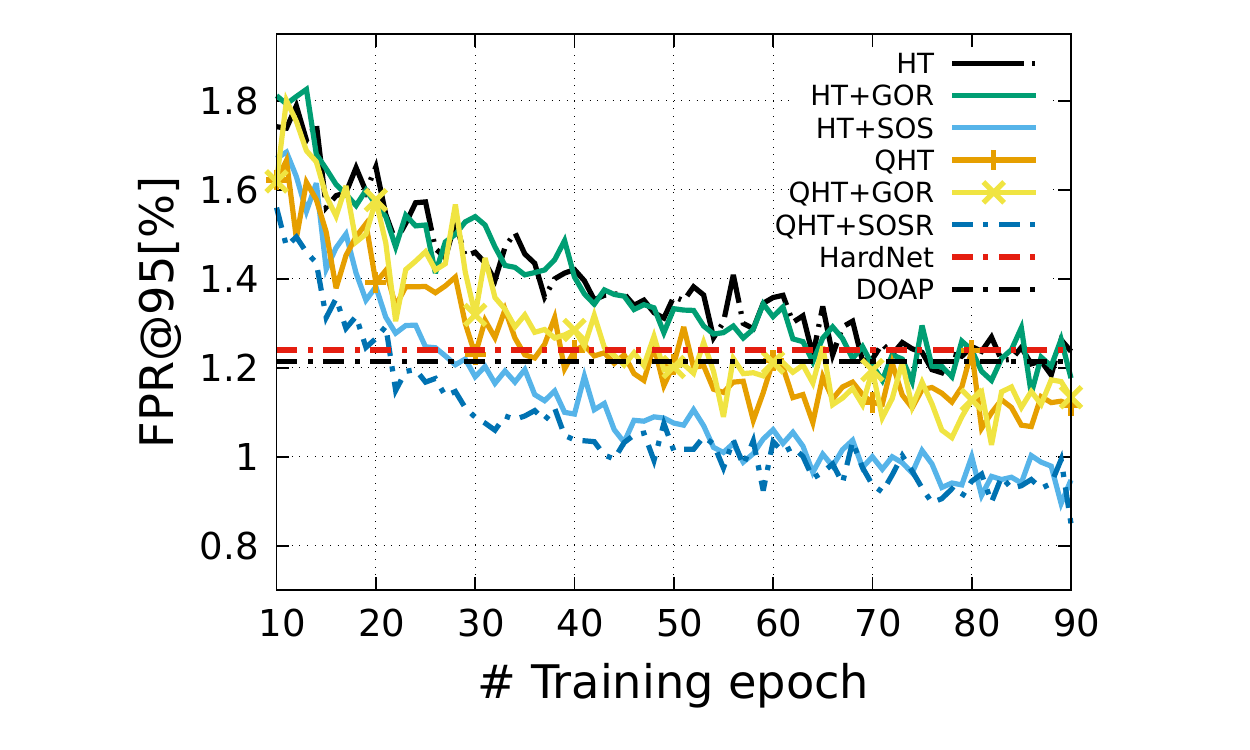}}
\end{minipage}
\begin{minipage}{0.325\linewidth}
\subfigure[Training with Adam.]{\label{fig:FRR_epoch_adam}{\includegraphics[trim={1.5cm 0.5cm 1.5cm 1.2cm},width=1\linewidth]{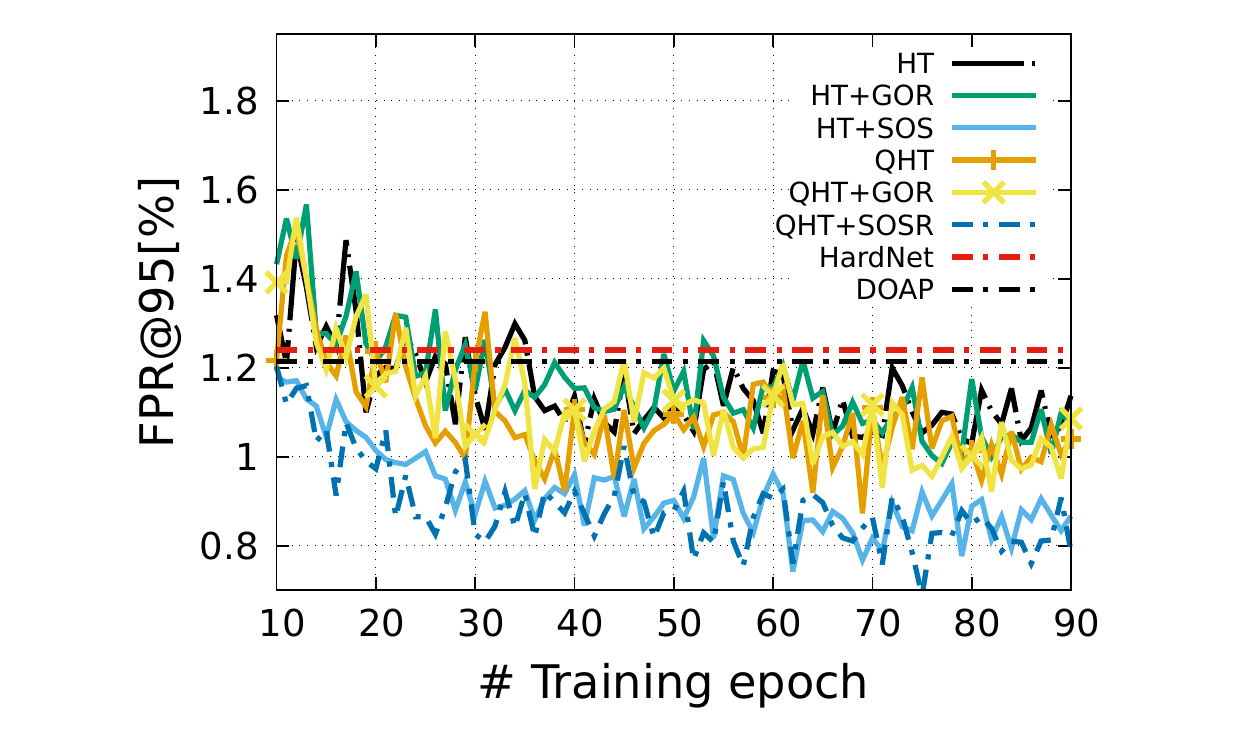}}}
\end{minipage}
\begin{minipage}{0.325\linewidth}
\subfigure[Impact of $N$ and $K$]{\label{fig:N_K}\includegraphics[trim={1.5cm 0.5cm 1.5cm 1.1cm},width=1\linewidth]{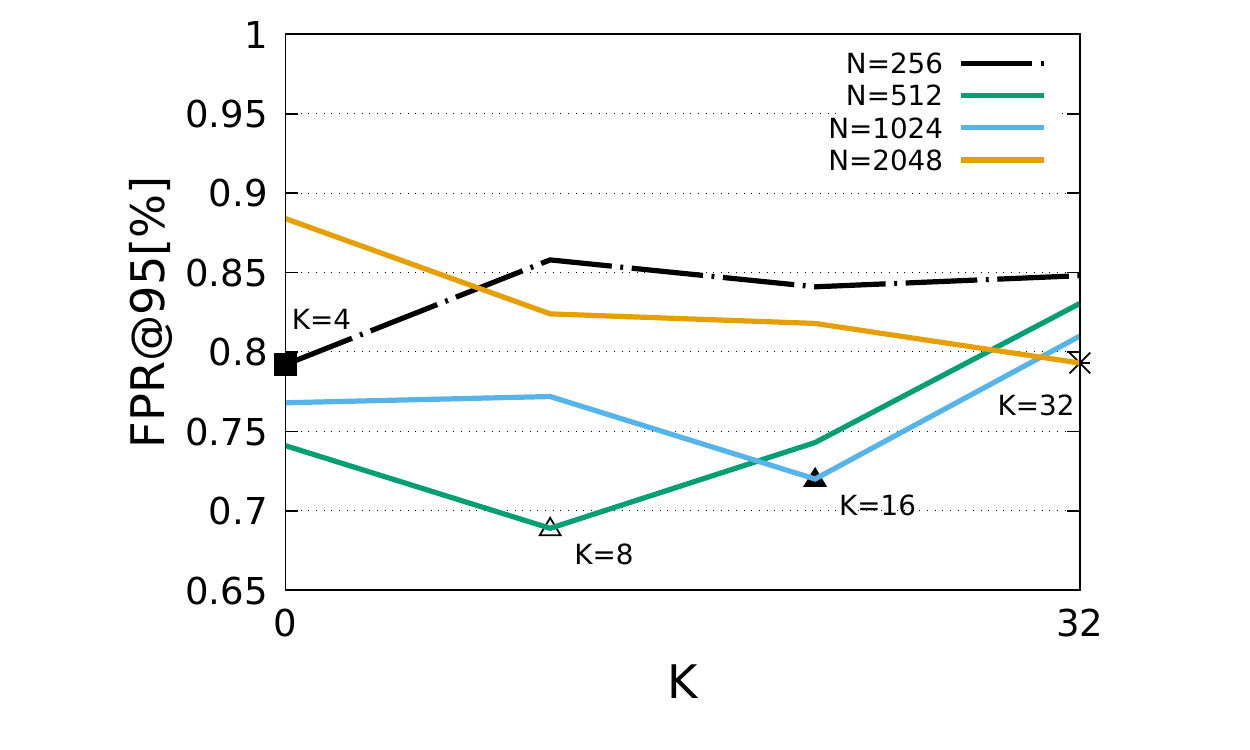}}
\end{minipage}
\caption{Analysis of Performance Improvements. HT stands for Hinge Triplet and QHT stands for Quadratic Hinge Triplet. All models are trained on Liberty and the FPR@95 is averaged over Yosemite and Notredame. Note that \descr is denoted as QHT+SOSR.}
\end{figure*}

First, we investigate the impact of different optimization methods, where the two most widely adopted methods, \ie, Stochastic Gradient Descent (SGD) and Adam~\cite{adam2014} are compared.
For SGD, we use a starting learning rate of $0.01$ and divided it by $10$ at epoch 50, and for Adam, we use the settings described in Sec.~\ref{sec:Experiments}.
As visible in Fig.~\ref{fig:FRR_epoch_sgd} and Fig.~\ref{fig:FRR_epoch_adam}, Adam~\cite{adam2014} leads to better performance compared to SGD.
Note that, using Hinge Triplet (HT) loss with Adam already surpasses the previous state-of-art method, \ie,  DOAP~\cite{doap2018} that uses a sophisticated ranking loss.

Second, we compare QHT against HT. As shown in Fig.~\ref{fig:FRR_epoch_sgd} and Fig.~\ref{fig:FRR_epoch_adam}, performance improvements from HT to QHT are quite obvious for both SGD and Adam cases. This is mainly due to the fact that QHT loss adaptively weights the gradients by the magnitude of the loss, \ie, $d_{\text{neg}} - d_{\text{pos}}$.

Third, we compare our \RegOurs with another regularization term recently proposed in ~\cite{spread_out}, \ie, the Global Orthogonal Regularization (GOR).
As shown in Fig.~\ref{fig:FRR_epoch_sgd} and Fig.~\ref{fig:FRR_epoch_adam}, \RegOurs achieves significant and consistent performance improvements across all training epochs, while the FPR curves with and without GOR are sometimes intertwined, showing minor performance enhancement, and this phenomenon is also observed in~\cite{doap2018}.

To sum up, Adam, QHT and \RegOurs bring on average $11.63$\%, $5.46$\%, and $\bm{19.49}$\% relative performance improvements, respectively.
Note that when calculating the relative performance increase caused by  \RegOurs, we average the FPR@95 over HT, QHT for both SGD and Adam from epoch $50$ to epoch $100$, with the same rule applying to Adam and QHT.

\begin{figure}
    \centering
    \includegraphics[trim={0cm 0cm 0cm 0cm},clip,width=\columnwidth]{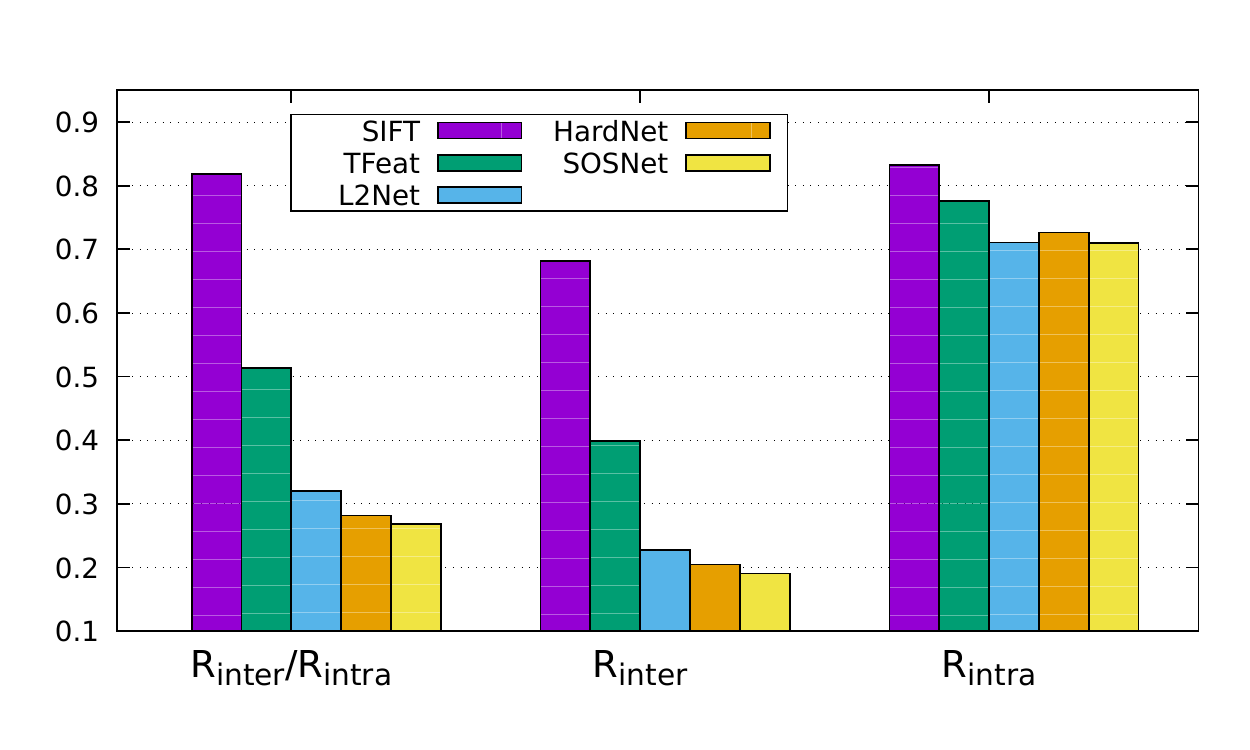}
    \caption{Performance in terms of the mean resultant length for the HPatches dataset. SIFT~\cite{sift2004} is normalized to have unit length.}
    \label{fig:R_trend}
\end{figure}

\subsection{Impact of $K$ and $N$}

As described in Sec.~\ref{sec:vMF Distribution}, each training batch is formed by $N$ pairs of patches, and within each batch, $K$ nearest neighbors are used to calculate \RegOurs.
In this section, we analyze the impact of the hyperparameters $N$ and $K$ on the matching performance \descr.
Specifically, we vary $N$ and $K$ from $256$ to $2048$ and $4$ to $32$, respectively.
All models are trained on Liberty and tested on the other two subsets, \ie, Notredame and Yosemite.
We report the mean FPR@95 of the two test sets in Fig.~\ref{fig:N_K}.
Across all the settings, $N=512$ with $K=8$ achieves the best performance.

\subsection{Analysis of the Descriptor Space}
\label{Analysis of the Descriptor Space}
To visualize the changes in the descriptor space caused by \RegOurs, we first conduct a toy experiment on the MNIST \cite{mnist1998} dataset. 
Specifically, we modify the L2Net architecture~\cite{l2net2017} by setting the number of output channels of the last convolutional layer as $3$.
The network is trained with a batch size of $20$, \ie, each batch contains $10$ classes with $2$ images per class. 
After training, we visualize the distribution of the descriptors on a unit sphere in Fig~\ref{fig:mnist_toy}.
It can be clearly seen that \RegOurs makes each cluster more concentrated, thus indicating that in the low dimensional space enforcing \sos constraint improves \fos.

Unlike clustering $10$ classes of images on a unit sphere, it is hard to directly visualize the distribution of descriptors on $\mathbb{S}^{127}$ from tens of thousands of classes. 
We have tried dimensionality reduction techniques such as tSNE~\cite{tsne2008}. However, it is hard to get any insightful conclusions visually about the structure of the descriptor space due to the distortions introduced by the dimensionality reduction process.

In order to provide quantitative results, we employ the evaluation method described in Sec.~\ref{sec:vMF Distribution}.
Specifically, we evaluate Eqn.~\eqref{eq:vmf_indicator} by using 90K randomly selected classes from the HPatches dataset.
To avoid noisy estimation of $R_{\text{inter}}$, like $R_{\text{intra}}$, we compute it by averaging 10K random tests, where in the $i^{\text{th}}$ test a $R^{\text{inter}}_i$ is estimated by sampling descriptors randomly from all classes(one descriptor per class).
The results are shown in Fig.~\ref{fig:R_trend}, and several interesting observations can be drawn:

\begin{itemize}
    \item
    The ratio $\rho$ drops in accordance with the performance ranking, \ie, SIFT $<$ TFeat $<$ L2Net $<$ HardNet $<$ \descr, indicating that it is a reasonable performance indicator.
    \item
    As performance increases from SIFT to \descr, $R_{\text{inter}}$ decreases monotonically, showing that the more space on the hypersphere has been exploited.
    Specifically, descriptors can harness more expressive power of $\mathbb{S}^{127}$ by exploiting more space of it, thus leading to better matching performance.
    \item
    With more space on $\mathbb{S}^{127}$ being exploited, $R_{\text{intra}}$ also drops, which means there are more scattered intra-class distributions. However, as long as the inter-class distribution is scattered enough, less concentrated intra-class distributions do not damage the matching performance.
    \item
    SIFT has the highest $R_{\text{intra}}$, indicating that the intra-class distributions are very concentrated. Meanwhile, it also has the highest $R_{\text{inter}}$, showing that most of the classes are gathered in a small region of $\mathbb{S}^{127}$, while leaving most of the area unexploited.

\end{itemize}

It is interesting to note that in low dimensional space (Fig.~\ref{fig:mnist_toy}), \RegOurs helps to make more concentrated intra-class distributions, while in high dimensional space (Fig.~\ref{fig:R_trend}) it helps to make inter-class distribution more scattered.
We argue that this phenomenon is related to the dimension of the descriptors.
When the descriptor space is of less complexity, \eg, $\mathbb{S}^{2}$, there is less flexibility for adjusting descriptor distributions. Therefore to ensure high \sosFullLower, \RegOurs enforces descriptors from the same class to become one point.
In contrast, for high dimensional descriptor space which is hard to visualize or even imagine, \eg, $\mathbb{S}^{127}$,  experimental results show that \RegOurs leads to a more scatter inter-class distribution, \ie, the descriptors exploit more area on the hypersphere.
To sum up, the adjustment of the descriptor space by \RegOurs leads to better matching performance, thus demonstrating our intuition that enforcing \sos in the training stage is reasonable. 

\section{Conclusions}
In this work, we propose a regularization term named \RegOursFull (\RegOurs) for the purpose of incorporating second order similarities into the leaning of local descriptors.
We achieve state-of-the-art performance on several standard benchmarks on different tasks including patch matching, verification, retrieval, and 3D reconstruction, thus demonstrating the effectiveness of the proposed \RegOurs.
Furthermore, we propose an evaluation method based on the von Mises-Fisher distribution to investigate the impact of enforcing second order similarity during training. By leveraging this evaluation method, we observe how the intra-class and inter-class distributions affect the performance of different descriptors.

\noindent \textbf{Acknowledgement}. This work is supported by the National Natural Science Foundation of China (61573352,61876180), the Young Elite Scientists Sponsorship Program by CAST (2018QNRC001), the Australian Research Council Centre of Excellence for Robotic Vision (project number CE140100016) and Scape Technologies.

{\small
\bibliographystyle{ieee}
\bibliography{egbib}
}

\end{document}